\documentclass[10pt,twocolumn,letterpaper]{article}

\usepackage{cvpr}              

\definecolor{cvprblue}{rgb}{0.21,0.49,0.74}
\usepackage[pagebackref,breaklinks,colorlinks,allcolors=cvprblue]{hyperref}

\def\paperID{2667} 
\def\confName{CVPR}
\def\confYear{2025}

\def\netName{BridgeAD}

\usepackage{makecell,pifont,multirow,multicol}
\usepackage{colortbl}
\usepackage{caption}
\usepackage{subcaption}
\usepackage{booktabs}
\usepackage{subcaption}
\usepackage{amssymb}
\usepackage{marvosym}
\usepackage{titletoc}
\usepackage{color}
\usepackage{xcolor}
\usepackage{multirow}
\usepackage{multicol}
\usepackage{pifont}
\usepackage{array}
\usepackage[accsupp]{axessibility} 

\title{Bridging Past and Future: End-to-End Autonomous Driving with Historical Prediction and Planning}

\author{
  Bozhou Zhang$^{1}$\quad\quad 
  Nan Song$^{1}$\quad\quad 
  Xin Jin$^{2}$\quad\quad
  Li Zhang$^{1}$\thanks{Corresponding author (\url{lizhangfd@fudan.edu.cn}).}
  \vspace{0.5em} 
  \\
  \quad\quad\quad \textsuperscript{1} School of Data Science, Fudan University \\
  \quad\quad\quad \textsuperscript{2} Eastern Institute of Technology 
  \vspace{.5em} 
  \\
  \url{https://github.com/fudan-zvg/BridgeAD}
}

\begin{document}


\maketitle
\begin{abstract}

End-to-end autonomous driving unifies tasks in a differentiable framework, enabling planning-oriented optimization and attracting growing attention.
Existing methods aggregate historical information either through dense historical bird’s-eye-view (BEV) features or by querying a sparse memory bank, following paradigms inherited from detection.
We argue that these paradigms either omit historical information in motion planning or fail to align with its multi-step nature, which requires predicting or planning multiple future time steps. 
In line with the philosophy of ``future is a continuation of past'', we propose~\textbf{\netName}, which reformulates motion and planning queries as multi-step queries to differentiate the queries for each future time step. This design enables the effective use of historical prediction and planning by applying them to the appropriate parts of the end-to-end system based on the time steps, which improves both perception and motion planning.
Specifically, historical queries for the current frame are combined with perception, while queries for future frames are integrated with motion planning. In this way, we bridge the gap between past and future by aggregating historical insights at every time step, enhancing the overall coherence and accuracy of the end-to-end autonomous driving pipeline.
Extensive experiments on the nuScenes dataset in both open-loop and closed-loop settings demonstrate that~\netName~achieves state-of-the-art performance.

\end{abstract}    
\section{Introduction}
\label{sec:introduction}

\begin{figure}[ht!]
    \centering
    
        \includegraphics[width=0.47\textwidth]
        {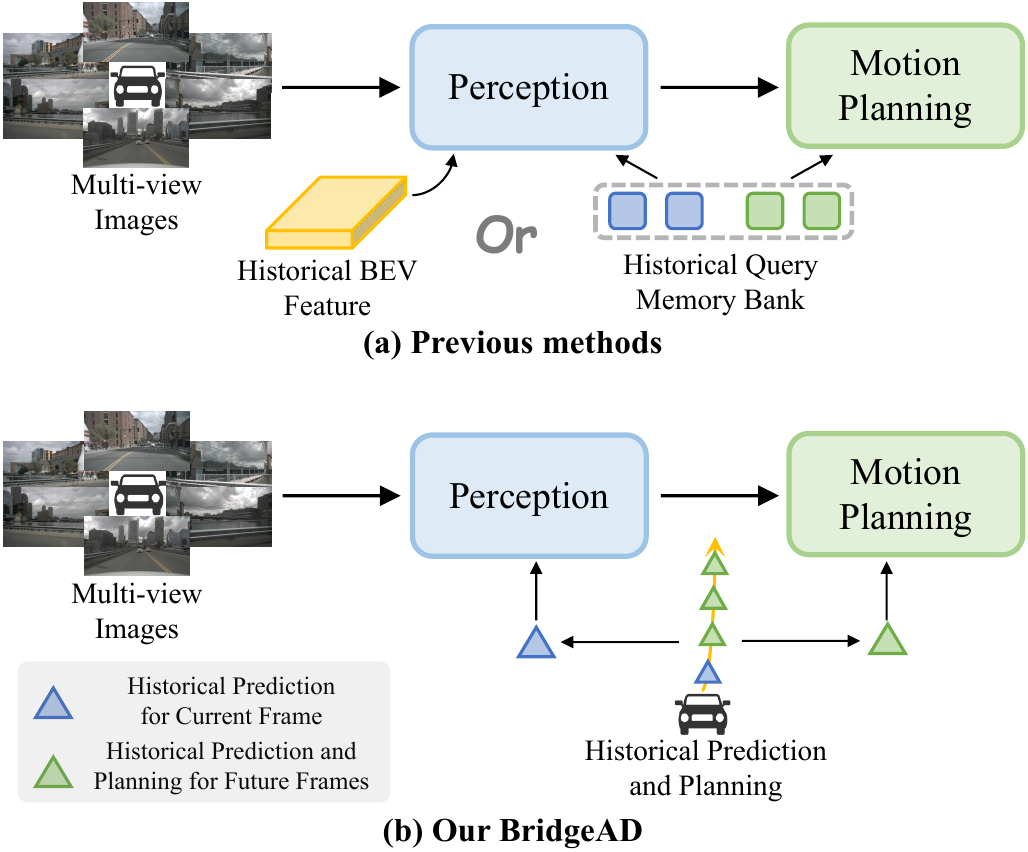}
        
    \caption{
    The primary distinction between previous methods and ours lies in how historical information is aggregated. As depicted in (a), previous methods either interact with historical BEV features within the perception module or utilize a historical query memory bank. As shown in (b), our~\netName~enhances end-to-end autonomous driving by incorporating {\bf historical prediction for the current frame} into the perception module and {\bf historical prediction and planning for future frames} into the motion planning module.
    }
    \label{fig:first}
\end{figure}

Autonomous driving~\cite{e2e} has progressed rapidly in recent years. Traditional systems use a modular approach, dividing tasks into perception~\cite{bevformer,streampetr,MapTR}, prediction~\cite{mtr,qcnet}, and planning~\cite{tuplan,plantf}, which simplifies each task but may interrupt the flow of information and lead to error accumulation. End-to-end methods~\cite{UniAD,vad} unify these tasks, enabling planning-oriented optimization and improved system coherence, and have gained increasing attention.

Current end-to-end methods largely originate from detection approaches~\cite{bevformer,streampetr,sparse4dv2}, adopting similar paradigms for utilizing temporal information to enhance performance. These paradigms are generally divided into two categories: dense methods~\cite{UniAD,vad}, which aggregate historical bird’s-eye-view (BEV) features, and sparse methods~\cite{sparsedrive,sparsead}, which rely on querying a sparse memory bank.
However, we argue that these paradigms are suboptimal. As shown in Figure~\ref{fig:first} (a), the former leverages temporal information solely in the perception module, overlooking its importance in motion planning. The latter performs a rough interaction with historical motion planning queries, where each query corresponds to a trajectory instance. This approach does not align with the multi-step nature of motion planning, which requires predicting or planning multiple future steps to account for varying agent states over time, leading to suboptimal results.

In this paper, we propose~{\bf \netName}, a framework to enhance end-to-end autonomous driving by leveraging historical prediction and planning, as shown in Figure~\ref{fig:first} (b). Embracing the idea that {\em future is a continuation of past}, we first decompose motion and planning queries into multi-step queries to address each future time step individually. Then, motion queries for the current frame, derived from historical prediction, are integrated into the perception module to enhance perception accuracy. Similarly, motion and planning queries for future frames, derived from historical prediction and planning, are combined within the motion planning module, allowing step-specific interactions to refine prediction and planning outcomes. Additionally, interactions between motion and planning queries at corresponding steps ensure consistency between the predictions of surrounding agents and the ego vehicle's planning across future time steps.
Through this design, we bridge past and future by merging historical prediction and planning with current perception and future motion planning. This approach enhances the entire end-to-end autonomous driving pipeline, creating a more cohesive system that improves the accuracy and consistency of perception, prediction, and planning across different time steps.

Our {\bf contributions} are summarized as follows: {\bf (i)} We represent motion and planning queries as multi-step queries, distinguishing each future time step to leverage historical insights at the step level. {\bf (ii)} We introduce~\netName, a novel framework that employs historical prediction and planning to enhance the end-to-end autonomous driving pipeline. {\bf (iii)} Extensive experiments on the nuScenes dataset, conducted in both open-loop and closed-loop settings, demonstrate that~\netName~achieves state-of-the-art performance.

\section{Related work}
\label{sec:relatedwork}

\paragraph{Perception.}
Perception extracts meaningful information from raw sensor data, primarily through 3D detection, multi-object tracking, and online mapping. For 3D detection, a series of approaches~\cite{bevdet4d,bevfusion} inspired by LSS~\cite{lss} obtain BEV representations from 2D image features using depth estimation; other approaches~\cite{bevformer,bevformerv2} use predefined BEV queries for feature sampling. Recent methods~\cite{streampetr,sparse4dv2} adopt a sparse approach, employing sparse queries for spatial-temporal aggregation.
For multi-object tracking (MOT), some methods~\cite{CenterPoint,streampetr} use the tracking-by-detection approach, while others~\cite{motr,mutr3d} employ track queries to continuously model tracked instances.
For online mapping, HDMapNet~\cite{hdmapnet} accomplishes this using BEV semantic segmentation with post-processing, while VectorMapNet~\cite{vectormapnet} employs a two-stage autoregressive transformer for vectorized map construction. MapTR~\cite{MapTR} and subsequent methods~\cite{maptrv2,maptracker} treat map elements as permutation-equivalent point sets, achieving impressive performance.

\paragraph{Motion prediction.}
Motion prediction aims to forecast agents' multi-modal future trajectories. Inspired by object queries in detection~\cite{detr}, some methods adopt a query-centric paradigm~\cite{mtr, qcnet, mtr++, gameformer,DeMo} to achieve strong performance in motion prediction benchmarks~\cite{WAYMO, argoverse2}. 
Some works aim to enhance motion prediction performance by incorporating historical predictions~\cite{hpnet} or employing a streaming approach~\cite{RealMotion}.
Other approaches~\cite{faf, pnpnet,intentnet} explore end-to-end motion prediction by first perceiving objects from multi-view images and then predicting their future trajectories. ViP3D~\cite{vip3d} leverages agent queries to jointly perform tracking and prediction, using images and HD maps as input.

\paragraph{Planning.}
Rule-based~\cite{tuplan,idm} and learning-based planners~\cite{plantf,pluto} are widely explored in planning benchmarks~\cite{nuplan}. 
Some works~\cite{gu2021belief,bouton2017belief,huang2024learning} explore the use of belief states to improve planning results or decision-making.
Recently, end-to-end planning has gained attention for its ability to integrate perception, prediction, and planning within a unified framework. 
Earlier approaches~\cite{TransFuser,endimitation} often bypass intermediate tasks such as perception and motion prediction. ST-P3~\cite{stp3} incorporates map perception, BEV occupancy prediction, and trajectory planning to derive ego vehicle planning results from surrounding camera views. Recently, UniAD~\cite{UniAD} has significantly advanced end-to-end autonomous driving by introducing a unified query design that integrates multiple tasks into a planning-oriented model, delivering impressive performance across perception, prediction, and planning.
VAD~\cite{vad} simplifies the pipeline by using vectorized map representations, achieving state-of-the-art planning performance with improved efficiency. GenAD~\cite{genad} adopts a generative framework that predicts the ego vehicle's future trajectories within a learned probabilistic latent space. SparseDrive~\cite{sparsedrive} employs a sparse scene representation and a parallel structure for its motion planner.
However, these methods do not fully explore how to leverage historical information to improve planning accuracy and continuity during continuous driving. Our~\netName~is the first to integrate this insight into its design.

\section{Methodology}
\label{sec:methodology}

\begin{figure*}[ht!]
    \centering
    
        \includegraphics[width=1\textwidth]
        {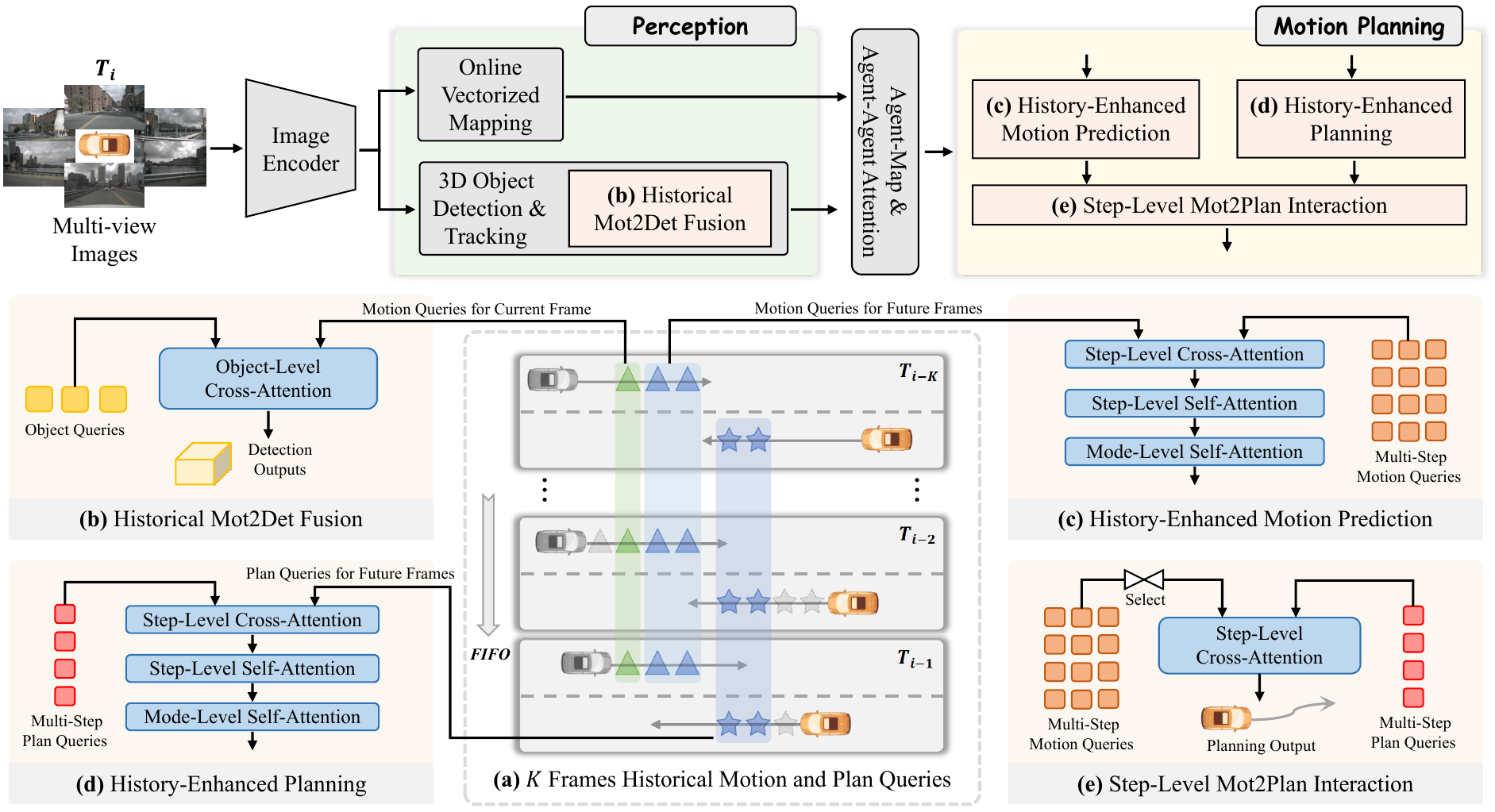}
        
    \caption{
    Overview of the~{\bf \netName}~framework: Multi-view images are first processed by the Image Encoder, after which both 3D objects and the vectorized map are perceived. (a)  The memory queue caches $K$ past frames of historical motion and planning queries. (b) The Historical Mot2Det Fusion Module is proposed to enhance detection and tracking by leveraging historical motion queries for the current frame. In the motion planning component, (c) the History-Enhanced Motion Prediction Module and (d) the History-Enhanced Planning Module aggregate multi-step historical motion and planning queries into queries for the future frames. Finally, (e) the Step-Level Mot2Plan Interaction Module facilitates interaction between multi-step motion queries and planning queries for corresponding future time steps.
    }
    \label{fig:main}
\end{figure*}

\subsection{Overview}
The framework of~\netName~is illustrated in Figure~\ref{fig:main}. It comprises three main components: image encoder, history-enhanced perception and history-enhanced motion planning. 
First, the image encoder extracts multi-scale spatial features from multi-view images. Next, the history-enhanced perception module employs a sparse approach for 3D object detection, tracking, and online vectorized mapping, integrating historical information through (b) the Historical Mot2Det Fusion Module, followed by agent-agent and agent-map attention. Finally, the history-enhanced motion planning module, consisting of (c) the History-Enhanced Motion Prediction Module, (d) the History-Enhanced Planning Module, and (e) the Step-Level Mot2Plan Interaction Module, generates motion prediction and planning outputs using historical data. Additionally, the memory queue (a) caches historical motion and planning queries to provide relevant historical information to the above modules.

\subsection{Multi-step motion and planning query caching}
The design of our~\netName~framework relies on a multi-step representation for motion and planning queries. Existing methods represent multi-modal motion queries as ${Q_{\rm mot}^{\rm previous}} \in \mathbb{R}^{N_{\rm a} \times M_{\rm mot} \times C}$, where $N_{\rm a}$, $M_{\rm mot}$, and $C$ denote the number of surrounding agents, the number of prediction modes, and the feature channels, respectively. Each query corresponds to a trajectory. 
In contrast, we define the motion queries as ${Q_{\rm mot}} \in \mathbb{R}^{N_{\rm a} \times M_{\rm mot} \times T_{\rm mot} \times C}$, where $T_{\rm mot}$ is the number of future time steps for prediction. Similarly, we represent planning queries as ${Q_{\rm plan}} \in \mathbb{R}^{M_{\rm plan} \times T_{\rm plan} \times C}$, where $M_{\rm plan}$ and $T_{\rm plan}$ denote the number of planning modes and future planning time steps, respectively.
In this way, we differentiate queries across time steps in motion planning, establishing the foundation for step-level interactions with historical information in subsequent modules. Motion and planning queries for the past $K$ frames are stored in a memory queue, which operates on a first-in, first-out (FIFO) basis: as new frame information is added, the oldest entry is removed, as illustrated in Figure~\ref{fig:main} (a).

\subsection{History-enhanced perception}

\paragraph{Detection, tracking, and online mapping.}
Given the multi-view images $I \in \mathbb{R}^{N_{\rm img} \times 3 \times H \times W}$, where $N_{\rm img}$ denotes the number of camera views, the image encoder~\cite{resnet} first extracts multi-view visual features, denoted as $\mathcal{F}$. These features are then used for perception.

The key components of perception are detection, tracking, and online mapping. We follow a sparse paradigm~\cite{streampetr,sparse4dv2,sparsedrive}. For detection, surrounding agents are represented by a set of object queries ${Q_{\rm obj}} \in \mathbb{R}^{N_{\rm a} \times C}$ and and anchor boxes ${B_{\rm obj}} \in \mathbb{R}^{N_{\rm a} \times 11}$, where each box is represented as $\{x,y,z,ln(w),ln(h),ln(l),sin(\theta),con(\theta),v_{x},v_{y},v_{z}\}$, containing location, dimensions, yaw angle, and velocity components, respectively. 
Several attention-based decoder layers~\cite{attention,deformableattn,flashattention} are used to refine the object queries and anchor boxes. These layers take the visual features $\mathcal{F}$, object queries $Q_{\rm obj}$, and anchor boxes $B_{\rm obj}$ as input and output classification scores along with anchor box offsets.
For tracking, we follow the ID assignment process in Lin \etal~\cite{sparse4dv3}, where each object query is assigned a unique ID.
For online mapping, we employ a vectorized representation~\cite{MapTR,vad,sparsedrive}, where map instances are represented as a set of map queries and points, utilizing a structure similar to that used in detection.

\paragraph{Historical Mot2Det fusion.}

As shown in Figure~\ref{fig:main} (b), the Historical Mot2Det Fusion Module aggregates historical prediction. As mentioned above, we extract the motion query corresponding to the current frame's time step from the cached historical motion queries over the past $K$ frames, yielding ${Q_{\rm m2d}} \in \mathbb{R}^{N_{\rm a} \times K \times C}$. An attention mechanism is then applied to interact between historical motion queries $Q_{\rm m2d}$ and object queries $Q_{\rm obj}$, as shown below:
\begin{equation}
Q_{\rm obj} = {\rm CrossAttn}({\rm Q} = Q_{\rm obj}, {\rm K, V} = Q_{\rm m2d}).
\end{equation}

Then, similar to the decoder layers in the detection module, classification scores along with anchor box offsets are output, and the refined object queries are passed to the following modules.

\subsection{History-enhanced motion planning}
After obtaining object and map queries from the perception module, the object queries and initialized ego query interact with map queries and each other via attention. These refined queries are then passed to the motion planning module, which predicts future trajectories for surrounding agents and plans the ego vehicle’s trajectory.

\paragraph{History-enhanced motion prediction.}

As shown above, we formulate motion queries as multi-step queries, ${Q_{\rm mot}} \in \mathbb{R}^{N_{\rm a} \times M_{\rm mot} \times T_{\rm mot} \times C}$, initialized from object queries. From the cached historical motion queries over the past $K$ frames, we extract the motion queries corresponding to the future 
$T_{\rm m2m}$ steps in each frame, yielding ${Q_{\rm m2m}} \in \mathbb{R}^{N_{\rm a} \times M_{\rm mot} \times K \times T_{\rm m2m} \times C}$. 
It is worth noting that $T_{\rm m2m}$ is smaller than the total time steps $T_{\rm mot}$ used for motion prediction, as historical data does not allow prediction as far into the future as required for the current frame.
Attention is then applied in three aspects: cross-attention between $Q_{\rm mot}$ and $Q_{\rm m2m}$, and self-attention on $Q_{\rm mot}$ at both the step and mode levels, as shown below:
\begin{equation}
\begin{split}
Q_{\rm mot} &= {\rm CrossAttn}({\rm Q} = Q_{\rm mot}, {\rm K, V} = Q_{\rm m2m}), \\
Q_{\rm mot} &= {\rm StepSelfAttn}(Q_{\rm mot}), \\
Q_{\rm mot} &= {\rm ModeSelfAttn}(Q_{\rm mot}).
\end{split}
\end{equation}

This process, illustrated in Figure~\ref{fig:main} (c), aggregates historical prediction information and enhances consistency across future time steps and trajectory modes.

\paragraph{History-enhanced planning.}
The planning module follows a similar process to the motion prediction module, as shown in Figure~\ref{fig:main} (d). Planning queries are initialized as multi-step queries, ${Q_{\rm plan}} \in \mathbb{R}^{M_{\rm plan} \times T_{\rm plan} \times C}$, from the ego query. Historical planning queries corresponding to the future $T_{\rm p2p}$ steps are extracted to form ${Q_{\rm p2p}} \in \mathbb{R}^{M_{\rm plan} \times K \times T_{\rm p2p} \times C}$. Similar to the motion prediction module, three types of attention are applied, as shown below:
\begin{equation}
\begin{split}
Q_{\rm plan} &= {\rm CrossAttn}({\rm Q} = Q_{\rm plan}, {\rm K, V} = Q_{\rm p2p}), \\
Q_{\rm plan} &= {\rm StepSelfAttn}(Q_{\rm plan}), \\
Q_{\rm plan} &= {\rm ModeSelfAttn}(Q_{\rm plan}).
\end{split}
\end{equation}

Notably, cross-attention in both the motion prediction and planning modules occurs between corresponding time steps. Specifically, historical motion queries interact with the $T_{\rm m2m}$ steps of all $T_{\rm mot}$ motion queries, and the same applies to the planning module. Historical information is then propagated to all steps of queries using two levels of self-attention.

\paragraph{Step-level Mot2Plan interaction.}

To improve consistency between motion prediction and planning, we introduce a module to interact motion queries and planning queries at the step level, as shown in Figure~\ref{fig:main} (e). Specifically, the $T_{\rm plan}$ steps of motion queries, representing the future states of surrounding agents within the planning time horizon, interact with the corresponding planning queries. 
We select the queries with the highest probability across $M_{\rm mot}$ modes based on the prediction scores to form ${Q_{\rm mot}^{*}} \in \mathbb{R}^{N_{\rm a} \times T_{\rm plan} \times C}$. The process is shown below:
\begin{equation}
\begin{split}
Q_{\rm mot}^{*} &= {\rm SelectWithScore}(Q_{\rm mot}), \\
Q_{\rm plan} &= {\rm CrossAttn}({\rm Q} = Q_{\rm plan}, {\rm K, V} = Q_{\rm mot}^{*}).
\end{split}
\end{equation}

Finally, the planning trajectories and scores are output. Following previous practice~\cite{UniAD,vad,sparsedrive}, we use three driving commands: turn left, turn right, and go straight, to select and obtain the final planning output.

\begin{table*} [ht!] 
    \centering

        \centering
        {\begin{tabular}[b]{l|c|cccc|cccc|c}
        \toprule[1.5pt]

        \multirow{2}{*}{\textbf{Method}} & \multirow{2}{*}{\textbf{Reference}} &
        \multicolumn{4}{c|}{\textbf{L2 ($m$)} $\downarrow$} & 
        \multicolumn{4}{c|}{\textbf{Col. Rate (\%)} $\downarrow$} 
        & \multirow{2}{*}{\textbf{FPS}}\\
        & & 1$s$ & 2$s$ & 3$s$ & Avg. & 1$s$ & 2$s$ & 3$s$ & Avg. & \\
        \midrule 
         
         OccWorld-D~\cite{occworld} & ECCV 2024 & 0.52 & 1.27 & 2.41 & 1.40 & 0.12 & 0.40 & 2.08 & 0.87 & - \\
         Drive-WM~\cite{Drivingintothefuture} & CVPR 2024 & 0.43 & 0.77 & 1.20 & 0.80 & 0.10 & 0.21 & 0.48 & 0.26 & - \\
         
         ST-P3~\cite{stp3} & ECCV 2022 & 1.33 & 2.11 & 2.90 & 2.11 & 0.23 & 0.62 & 1.27 & 0.71 & 1.6 \\
         GenAD~\cite{genad} & ECCV 2024 & 0.36 & 0.83 & 1.55 & 0.91 & 0.06 & 0.23 & 1.00 & 0.43 & \bf 6.7 \\   
         UniAD$^\dagger$~\cite{UniAD} & CVPR 2023 & 0.45 & 0.70 & 1.04 & 0.73 & 0.62 & 0.58 & 0.63 & 0.61 & 1.8 \\
         VAD$^\dagger$~\cite{vad} & ICCV 2023 & 0.41 & 0.70 & 1.05 & 0.72 & 0.03 & 0.19 & 0.43 & 0.21 & 4.5 \\
         SparseDrive$^\dagger$~\cite{sparsedrive} & arXiv 2024 & 0.30 & 0.58 & 0.95 & 0.61 & \bf 0.01 & \bf 0.05 & 0.23 & 0.10 & 6.1 \\
         \rowcolor{gray!20}
         \netName-S~(Ours)~& - &  \bf 0.29 & \bf 0.57 & \bf 0.92 & \bf 0.59 & \bf 0.01 & \bf 0.05 & \bf 0.22 & \bf 0.09 & 5.0 \\
         \rowcolor{gray!20}
         \netName-B~(Ours)~& - & \bf 0.28 & \bf 0.55 & \bf 0.92 & \bf 0.58 & \bf 0.00 & \bf 0.04 & \bf 0.20 & \bf 0.08 & 3.9 \\
         
         \bottomrule[1.5pt]
        \end{tabular}}
    
    \caption{\textbf{Open-loop planning results} on the nuScenes~\cite{nuscenes} validation dataset. $\dag$ indicates evaluation with the official checkpoint. FPS is measured on one NVIDIA RTX 3090 GPU with batch size 1, while UniAD's is on one NVIDIA Tesla A100. To avoid the ego-status leakage problem, as proposed by Li \etal~\cite{bevplanner}, we do not use the ego status as input.}
    \label{tab:openloop}
\end{table*}
\begin{table} [ht!] 
    \centering

        \centering
        \resizebox{1.0\columnwidth}{!}
        {\begin{tabular}[b]{l|c|cc}
        \toprule[1.5pt]

        \textbf{Method} & \textbf{Post-proc.} & \textbf{Score} $\uparrow$ & \textbf{Col. Rate~(\%)} $\downarrow$ \\ 
        \midrule 
         
         VAD~\cite{vad} & \ding{55} & 0.66 & 92.5 \\ 
         UniAD~\cite{UniAD} & \ding{55} & 0.73 & 88.6 \\ 
         SparseDrive$^{\dag}$~\cite{sparsedrive}  & \ding{55} & 0.92 & 93.9 \\ 
         \rowcolor{gray!20}
         \netName-S~(Ours)~& \ding{55} & \bf 1.52 & \bf 76.2 \\
         \rowcolor{gray!20}
         \netName-B~(Ours)~& \ding{55} & \bf 1.60 & \bf 72.6 \\\midrule
         
         VAD~\cite{vad} & \ding{51} & 2.75 & 50.7 \\ 
         UniAD~\cite{UniAD} & \ding{51} & 1.84 & 68.7 \\
         \rowcolor{gray!20}
         \netName-S~(Ours)~& \ding{51} & \bf 2.98 & \bf 46.1 \\
         \rowcolor{gray!20}
         \netName-B~(Ours)~& \ding{51} & \bf 3.06 & \bf 44.3 \\
         \bottomrule[1.5pt]
        \end{tabular}}
    
    \caption{\textbf{Closed-loop simulation results} on nuScenes dataset with NeuroNCAP~\cite{neuroncap} benchmark. $\dag$ indicates evaluation with official checkpoint. ``Post-proc." refers to trajectory post-processing, as proposed in UniAD.}
    \label{tab:closeloop}
\end{table}
\begin{table} [ht!] 
    \centering

        \centering
        \resizebox{1.0\columnwidth}{!}
        {\begin{tabular}[b]{l|c|c
        |c|c}
        \toprule[1.5pt]

        \multirow{2}{*}{\textbf{Method}} & \textbf{ADE~($m$)} $\downarrow$ & \textbf{FDE~($m$)} $\downarrow$ & \textbf{MR} $\downarrow$ & \textbf{EPA} $\uparrow$ \\ 
        & Car / Ped & Car / Ped & Car / Ped & Car / Ped\\\midrule 
         
         ViP3D~\cite{vip3d} & 2.05 / - & 2.84 / - & 0.25 / - & 0.23 / - \\
         UniAD~\cite{UniAD} & 0.71 / 0.78 & 1.02 / 1.05 & 0.15 / {\bf 0.12} & 0.46 / 0.35 \\ 
         SparseDrive~\cite{sparsedrive}  & {\bf 0.62} / 0.72 & 0.99 / 1.07 & 0.14 / 0.14 & 0.48 / 0.41 \\ 
         \rowcolor{gray!20}
         \netName-S~(Ours)~& \bf 0.62 / \bf 0.70 & \bf 0.98 / \bf 0.99 & \bf 0.13 / \bf 0.13 & \bf 0.50 / \bf 0.44 \\
         \rowcolor{gray!20}
         \netName-B~(Ours)~& \bf 0.60 / \bf 0.70 & \bf 0.96 / \bf 0.98 & \bf 0.13 / \bf 0.12 & \bf 0.52 / \bf 0.45 \\
         \bottomrule[1.5pt]
        \end{tabular}}
    
    \caption{Comparison of \textbf{motion prediction results} of state-of-the-art methods. We evaluate two main categories: cars and pedestrians.}
    \label{tab:motion}
\end{table}
\begin{table} [ht!] 
    \centering
    \begin{subtable}{0.45\textwidth}
        \centering
        \resizebox{0.80\columnwidth}{!}{\begin{tabular}[b]{l|c|c|c}
        \toprule[1.5pt]

        \textbf{Method} & \textbf{Backbone} & \textbf{mAP} $\uparrow$ & \textbf{NDS} $\uparrow$ \\\midrule 
        
         VAD$^{\dag}$~\cite{vad}  & R50  & 0.273 & 0.397 \\
         SparseDrive~\cite{sparsedrive} & R50 & 0.418 & 0.525 \\
         \rowcolor{gray!20}
         \netName-S~(Ours)~& R50 & \bf 0.423 & \bf 0.534 \\\midrule
         BEVFormer~\cite{bevformer} & R101 & 0.416 & 0.517 \\ 
         UniAD~\cite{UniAD} & R101 & 0.380 & 0.498 \\  

         \rowcolor{gray!20}
         \netName-B~(Ours)~& R101 & \bf 0.507 & \bf 0.594 \\
         \bottomrule[1.5pt]
        \end{tabular}}
    
    \caption{3D detection results.}
    \end{subtable}

    \vspace{0.6em}
    
    \begin{subtable}{0.47\textwidth}
        \centering
        \resizebox{1\columnwidth}{!}{\begin{tabular}[b]{l|c|c|c|c}
        \toprule[1.5pt]

        \textbf{Method} & \textbf{Backbone} & \textbf{AMOTA} $\uparrow$ & \textbf{AMOTP} $\downarrow$ & \textbf{IDS} $\downarrow$ \\\midrule 

         ViP3D~\cite{vip3d} & R50 & 0.217 & 1.625 & - \\
         SparseDrive~\cite{sparsedrive} & R50 & 0.386 & 1.254 & 886 \\
         \rowcolor{gray!20}
         \netName-S~(Ours)~& R50 & \bf 0.398 & \bf 1.232 & \bf 639 \\\midrule 
         UniAD~\cite{UniAD} & R101 & 0.359 & 1.320 & 906 \\

         \rowcolor{gray!20}
         \netName-B~(Ours)~& R101 & \bf 0.512 & \bf 1.080 & \bf 544 \\
         \bottomrule[1.5pt]
        \end{tabular}}
    
    \caption{Multi-object tracking results.}
    \end{subtable}

    \vspace{0.6em}

    \caption{Comparison of \textbf{perception results} of state-of-the-art perception or end-to-end methods. $\dag$ indicates evaluation with official checkpoint.}
    \label{tab:dettrackmap}
\end{table}

\subsection{End-to-end learning}
The loss functions consist of four tasks: detection ($\mathcal{L}_{det}$), online mapping ($\mathcal{L}_{map}$), motion prediction ($\mathcal{L}_{mot}$), and planning ($\mathcal{L}_{plan}$). The loss for each task is divided into regression and classification components. For regression, we use L1 loss, and for classification, we use Focal loss~\cite{focalloss}. For the multi-modal motion prediction and planning tasks, we apply a winner-takes-all strategy.
The overall loss function for end-to-end training is as follows:
\begin{equation}
\mathcal{L}_{total}=\mathcal{L}_{det}+\mathcal{L}_{map}+\mathcal{L}_{mot}+\mathcal{L}_{plan}.
\end{equation}

Further details of the model and loss function are provided in the supplementary materials.

\section{Experiments}
\label{sec:experiment}

\subsection{Experimental settings}
\paragraph{Datasets and evaluation metrics.}
We conduct our experiments on the challenging nuScenes~\cite{nuscenes} dataset, which comprises 1,000 driving scenes, each lasting 20 seconds. The dataset provides semantic maps and 3D object detection annotations for keyframes, with samples annotated at 2Hz, including six camera images per keyframe. We perform open-loop testing on nuScenes following previous work~\cite{UniAD,vad} and conduct closed-loop testing in the NeuroNCAP~\cite{neuroncap} simulator based on nuScenes. NeuroNCAP is a photorealistic closed-loop simulation framework providing diverse safety-critical scenarios recorded from nuScenes, which are not feasible to collect in the real world.
For open-loop evaluation, we use the L2 Displacement Error metric, consistent with VAD~\cite{vad}, and the Collision Rate~\cite{stp3,bevplanner} as defined in~\cite{bevplanner,sparsedrive}. For closed-loop evaluation, we apply the NeuroNCAP Score and Collision Rate~\cite{neuroncap}.
Additional metrics for perception and prediction tasks are consistent with previous work~\cite{UniAD}. Further details are provided in the supplementary materials.

\paragraph{Implementation details.} 
\netName~plans a 3-second future trajectory for the ego vehicle and forecasts a 6-second future trajectory for surrounding agents. This setup results in a motion prediction time step, $T_{\rm mot}$, of 12 and a planning time step, $T_{\rm plan}$, of 6. 
The historical time steps for motion prediction, $T_{\rm m2m}$, is set to 6, and for planning, $T_{\rm p2p}$ is set to 3. We cache the past $K=3$ frames of motion and planning queries in the memory queue.
Our~\netName~model has two variants:~\netName-S and~\netName-B. For~\netName-S, ResNet50~\cite{resnet} is used as the backbone network to encode image features, with an input image size of $256\times704$; this is our default model. For~\netName-B, ResNet101 is used with an input image size of $512\times1408$. 
In training, we use the AdamW~\cite{adamw} optimizer with Cosine Annealing~\cite{cosineanneal}, a weight decay of $1\times10$$^{-3}$, and an initial learning rate of $1\times10$$^{-4}$. Training is conducted in two stages: one focused on perception tasks and the other on end-to-end training. Experiments are conducted on 8 NVIDIA RTX A6000 GPUs. Additional configuration details and further experiments are provided in the supplementary materials.

\subsection{Comparison with state of the art}
\paragraph{Open-loop planning results.}
As shown in Table~\ref{tab:openloop}, we compare the open-loop planning performance of our~\netName~ with recent top-performing methods, including both end-to-end autonomous driving~\cite{UniAD,vad,sparsedrive,genad} and world model~\cite{occworld,Drivingintothefuture} approaches. Our~\netName~achieves state-of-the-art performance. To address the issue raised by Li \etal~\cite{bevplanner} regarding over-reliance on ego vehicle status for future path planning, our~\netName~does not use ego status as input. Despite this, our method outperforms others that do rely on ego status.


\paragraph{Closed-loop planning results.}
We adopt~\netName~for closed-loop evaluation using the NeuroNCAP~\cite{neuroncap} simulator based on the nuScenes~\cite{nuscenes} dataset, which provides photorealistic, safety-critical scenarios for testing. Our~\netName~achieves significantly better performance than previous methods~\cite{UniAD,vad,sparsedrive}, with or without the trajectory post-processing proposed by UniAD~\cite{UniAD}, as shown in Table~\ref{tab:closeloop}. 
Specifically, without post-processing, the NeuroNCAP score of our~\netName-S~is {\bf 65\%} higher than SparseDrive and reduces the collision rate by {\bf 12.4\%} compared to UniAD.
The results demonstrate that our model improves continuity and consistency in planning across continuous driving scenes by effectively aggregating historical information, highlighting the potential of~\netName~in closed-loop simulation. In contrast, other methods either neglect historical information in motion planning~\cite{UniAD,vad} or fail to effectively incorporate it at the current frame~\cite{sparsedrive}, which allows them to perceive surrounding agents but limits their ability to avoid collisions. We highlight this advantage in the qualitative analysis section to further support the insights of our method.

\paragraph{Perception and motion prediction results.}
The perception results are shown in Table~\ref{tab:dettrackmap}, and the motion prediction results in Table~\ref{tab:motion}. By leveraging historical information and multi-step motion query representation, our~\netName~achieves superior performance across all metrics compared to other methods. Similar improvements are also evident in the perception results, for both detection and tracking.

\begin{table*} [ht!] 
    \centering

        \centering
        {\begin{tabular}[b]{c|cc|cccc|cccc}
        \toprule[1.5pt]

        \multirow{2}{*}{\textbf{ID}} &
        \multirow{2}{*}{\textbf{HisPlan}} &
        \multirow{2}{*}{\textbf{Mot2Plan}} & 
        \multicolumn{4}{c|}{\textbf{L2 ($m$)} $\downarrow$} & 
        \multicolumn{4}{c}{\textbf{Col. Rate (\%)} $\downarrow$}  \\
        & & & 1$s$ & 2$s$ & 3$s$ & Avg. & 1$s$ & 2$s$ & 3$s$ & Avg. \\
        \midrule 
         
         1 &  & \ding{51} &  0.35 & 0.68 & 1.10 & 0.71 & 0.01 & 0.11 & 0.34 & 0.15 \\
         2 & \ding{51} &  &  0.33 & 0.65 & 1.07 & 0.68 & 0.01 & 0.13 & 0.40 & 0.18 \\
         \rowcolor{gray!20}
         3 &\ding{51} & \ding{51} &  0.29 & 0.57 & 0.92 & 0.59 & 0.01 & 0.05 & 0.22 & 0.09\\
         
         \bottomrule[1.5pt]
        \end{tabular}}
    
    \caption{Ablation study on the History-Enhanced Planning module and Step-Level Mot2Plan Interaction module.}
    \label{tab:abl_main}
\end{table*}
\begin{table*} [ht!] 
    \centering

        \centering
        {\begin{tabular}[b]{c|cc|cc|cc|ccc}
        \toprule[1.5pt]

        \multirow{2}{*}{\textbf{ID}} & \multirow{2}{*}{\textbf{Mot2Det}} &
        \multirow{2}{*}{\textbf{HisMot}} &
        \multicolumn{2}{c|}{\textbf{Detection}} & \multicolumn{2}{c|}{\textbf{Tracking}} & \multicolumn{3}{c}{\textbf{Motion Prediction}}\\
         & &  & \textbf{mAP} $\uparrow$ & \textbf{NDS} $\uparrow$ & \textbf{AMOTA} $\uparrow$ & \textbf{AMOTP} $\downarrow$ & \textbf{ADE~($m$)} $\downarrow$ & \textbf{FDE~($m$)} $\downarrow$ & \textbf{EPA} $\uparrow$ \\
        \midrule 
         
         1 & \ding{51} &  & 0.412 & 0.526 & 0.387 & 1.240 & 0.66 / 0.75 & 1.05 / 1.08 & 0.47 / 0.40\\
         2 & & \ding{51} & 0.404 & 0.512 & 0.369 & 1.260 & 0.62 / 0.69 & 0.99 / 0.98 & 0.49 / 0.43 \\
         \rowcolor{gray!20}
         3 &  \ding{51} & \ding{51} & 0.423 & 0.534 & 0.398 & 1.232 & 0.62 / 0.70 & 0.98 / 0.99 & 0.50 / 0.44 \\
         
         \bottomrule[1.5pt]
        \end{tabular}}
    
    \caption{Ablation study on the Historical Mot2Det Fusion module and History-Enhanced Motion Prediction module. We evaluate motion prediction for cars and pedestrians.}
    \label{tab:abl_detmot}
\end{table*}

\begin{table} [t!] 
    \centering

        \centering
        \resizebox{1.0\columnwidth}{!}
        {\begin{tabular}[b]{cc|c|c}
        \toprule[1.5pt]

        \textbf{SLA} &
        \textbf{MLA} &
        Avg. \textbf{L2 ($m$)} $\downarrow$  & 
        Avg. \textbf{Col. Rate (\%)} $\downarrow$ \\
        \midrule 
        
         \ding{51} &  & 0.66 & 0.17 \\
          & \ding{51} & 0.64 & 0.15 \\
         \rowcolor{gray!20}
         \ding{51} & \ding{51} & 0.59 & 0.09 \\
         \bottomrule[1.5pt]
        \end{tabular}}
    
    \caption{Ablation study on step-level self-attention (SLA) and mode-level self-attention (MLA).}
    \label{tab:abl_attn}
\end{table}
\begin{table} [t!] 
    \centering

        \centering
        \resizebox{1.0\columnwidth}{!}
        {\begin{tabular}[b]{cc|c|c}
        \toprule[1.5pt]

        \textbf{HisMot} &
        \textbf{HisPlan} &
        Avg. \textbf{L2 ($m$)} $\downarrow$  & 
        Avg. \textbf{Col. Rate (\%)} $\downarrow$ \\
        \midrule 
        
         5 & 3 & 0.63 & 0.13 \\
         7 & 3 & 0.62 & 0.09 \\\midrule
         6 & 2 & 0.64 & 0.13 \\
         6 & 4 & 0.60 & 0.11 \\\midrule
         \rowcolor{gray!20}
         6 & 3 & 0.59 & 0.09 \\
         \bottomrule[1.5pt]
        \end{tabular}}
    
    \caption{Ablation study on the number of time steps for aggregating historical information.}
    \label{tab:abl_state}
\end{table}

\subsection{Ablation study}

\paragraph{Effects of designs for planning.}
We conduct experiments on our planning design, as shown in Table~\ref{tab:abl_main}. In ID-1, we remove the History-Enhanced Planning module, and in ID-2, we remove the Step-Level Mot2Plan Interaction module. The results show that removing either module leads to a significant reduction in planning performance compared to~\netName~in ID-3. This demonstrates that historical planning information and prediction of surrounding agents play a crucial role in improving ego vehicle planning.

\paragraph{Effects of designs for perception and prediction.} 
We conduct experiments on our design for perception and prediction, with results shown in Table~\ref{tab:abl_detmot}. In ID-1, when the History-Enhanced Motion Prediction module is removed, motion prediction performance significantly declines. Although the Historical Mot2Det Fusion module is used, detection and tracking performance does not match that in ID-3 due to suboptimal prediction. Similarly, in ID-2, detection and tracking performance significantly decreases without the Historical Mot2Det Fusion module.

\begin{figure}[t!]
    \centering
    
        \includegraphics[width=0.47\textwidth]{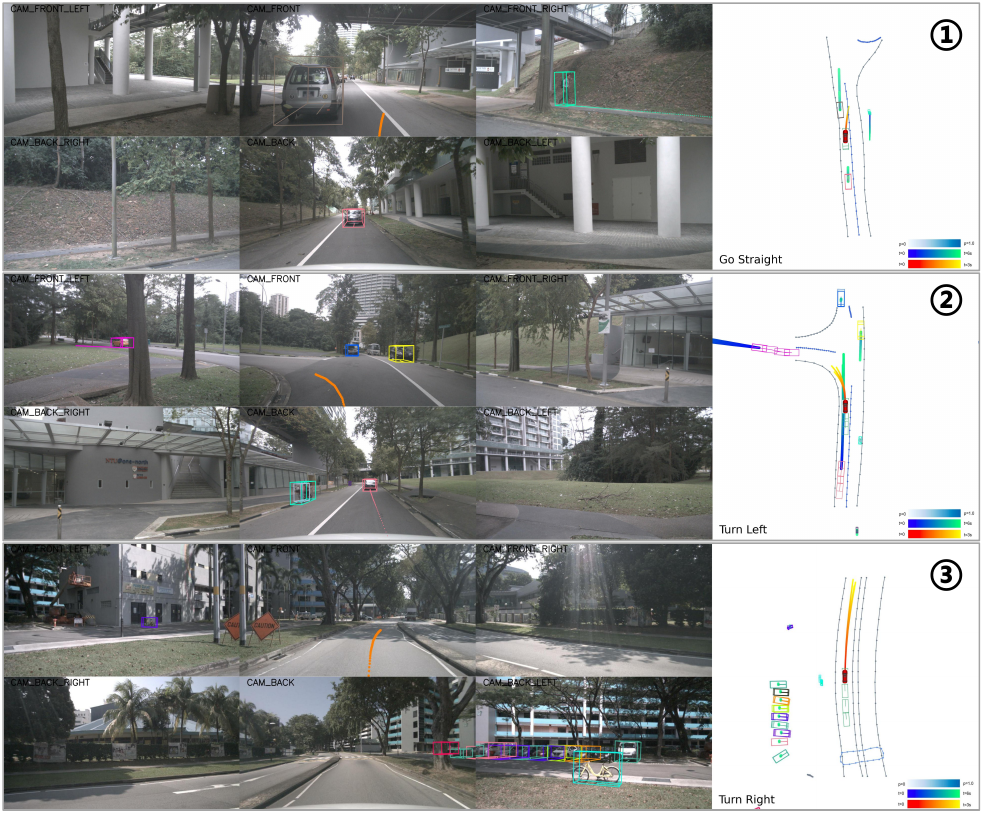}
        
    \caption{
    Qualitative results in the open-loop evaluation show that our~\netName~accurately produces planning outputs.
    }
    \label{fig:open}
\end{figure}
\begin{figure*}[t!]
    \centering

        \includegraphics[width=1\textwidth]{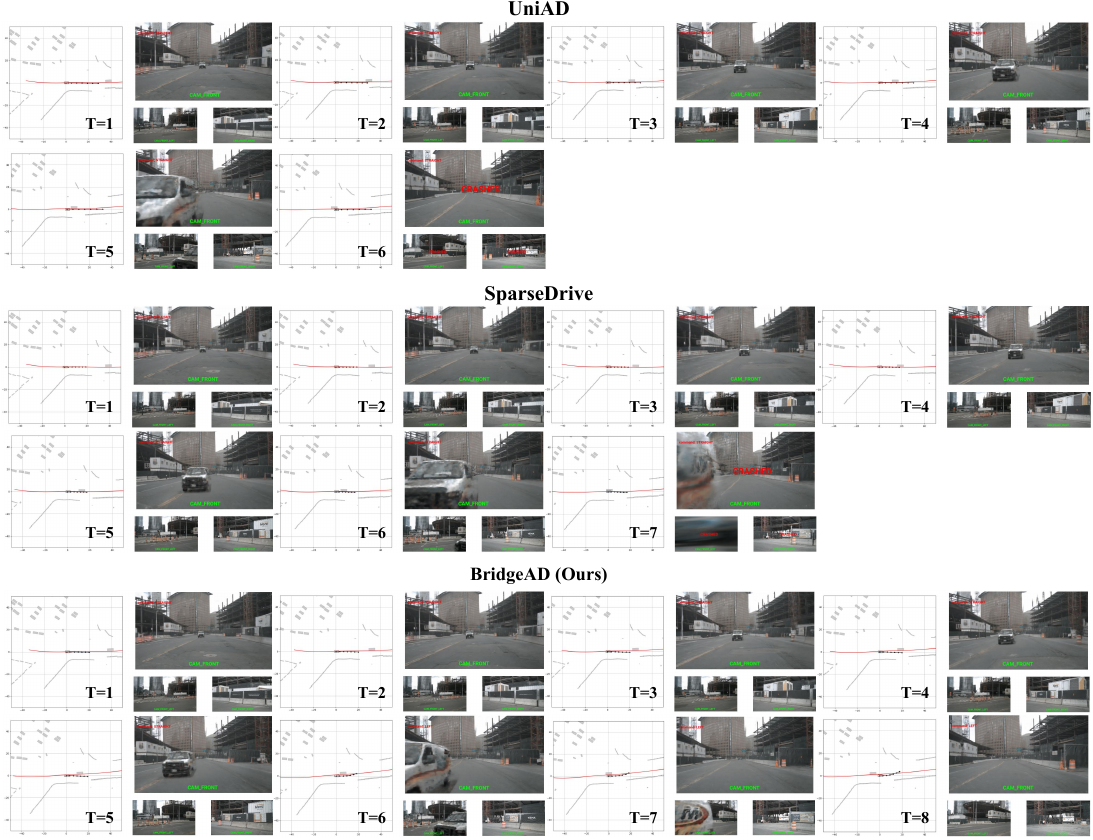}

    \caption{
    Qualitative results in the closed-loop evaluation demonstrate that our~\netName~effectively avoids collisions in safety-critical scenarios.
    }
    \label{fig:close}
\end{figure*}

\paragraph{Effects of self-attention in planning.}
We conduct an ablation study on the effect of step-level self-attention and mode-level self-attention in the planning module, as shown in Table~\ref{tab:abl_attn}. The results show that without either type of self-attention, planning performance significantly decreases. Without self-attention, historical information can only be aggregated for the $T_{\rm p2p}$ steps of planning queries. The step-level and mode-level self-attention mechanisms propagate this information across all planning steps and modes, enhancing both the accuracy and consistency of planning at each time step.

\paragraph{Effects of the number of time steps in aggregating historical information.}
As shown in Table~\ref{tab:abl_state}, we investigate performance variations based on the number of time steps used for aggregating historical information in motion and planning queries. 
We fix the time steps for aggregating historical information in planning queries at 3, while varying them in motion queries (upper part). Similarly, we fix the time steps in motion queries at 6, while varying them in planning queries (lower part).
We observe that the best results are achieved when the number of time steps for interacting with historical information is 6 for motion queries and 3 for planning queries.

\subsection{Efficiency analysis}
As shown in Table~\ref{tab:openloop}, we compare the Frames Per Second (FPS) of our~\netName~and other end-to-end methods. FPS for all models, except UniAD, is measured on a single NVIDIA RTX 3090 GPU with a batch size of 1. For UniAD, we use the official FPS value, measured on an NVIDIA Tesla A100 GPU. Our~\netName~achieves high performance with reasonable efficiency. The inference latency of our model is 157.2 ms, significantly faster than VAD's 224.3 ms and UniAD's 555.6 ms.

\subsection{Qualitative analysis}
\vspace{0.3em}
We present qualitative results of open-loop and closed-loop evaluation on the nuScenes~\cite{nuscenes} dataset. As shown in Figure~\ref{fig:open}, we display the perception and prediction outcomes along with the planning of the ego vehicle in both surrounding images and the Bird’s Eye View (BEV) in the open-loop setting. In Figure~\ref{fig:close}, we illustrate the closed-loop simulation results for a safety-critical scenario in which our~\netName~model successfully avoids a collision with an oncoming vehicle traveling in the wrong direction by steering appropriately. In contrast, UniAD~\cite{UniAD} and SparseDrive~\cite{sparsedrive} either fail to steer or do not steer sufficiently, resulting in a crash. 
Qualitative results for the closed-loop simulation show that our model, by aggregating historical motion and planning information, forms a continuous understanding of nearby vehicles’ motions, enabling coherent driving actions that successfully avoid collisions with oncoming vehicles.
Additional qualitative results and failure cases are provided in the supplementary materials.

\section{Conclusion}
\label{sec:conclusion}
In this paper, we propose~\netName, an end-to-end framework that enhances autonomous driving by integrating historical prediction and planning across perception, prediction, and planning stages. By representing motion and planning queries as multi-step queries, we enable step-specific interactions and leverage temporal information to improve coherence across future time steps. Extensive experiments on the nuScenes dataset in both open-loop and closed-loop scenarios demonstrate that~\netName~achieves superior performance. Our results highlight the potential of incorporating historical insights to bridge past and future, advancing technological progress in autonomous driving.

\section*{Acknowledgments}
This work was supported in part by National Natural Science Foundation of China (Grant No. 62376060).



{
    \small
    \bibliographystyle{ieeenat_fullname}
    \bibliography{main}
}


\maketitlesupplementary

\startcontents
{
    \hypersetup{linkcolor=black}
    \printcontents{}{1}{}
}
\newpage


\section{Methodology}
\subsection{Model details}
The perception component of our model follows a sparse paradigm~\cite{sparse4dv2,sparse4dv3,sparsedrive}. For detection, after obtaining the initialized object queries $Q_{\rm obj}$ and multi-view visual features $\mathcal{F}$, several decoder layers are applied. These layers include attention mechanisms across object queries, deformable aggregation with visual features, and a feedforward network~\cite{sparse4dv2}. 
The Historical Mot2Det Fusion Module, designed by us, follows the above modules to refine the object queries and detection outputs using historical prediction.
For online mapping, the structure is similar to that used in detection~\cite{sparsedrive}.
For multi-head attention, Flash Attention~\cite{flashattention} is adopted to save GPU memory.

\subsection{Loss function}
As stated in the End-to-End Learning section of the main paper, the loss function for each task is divided into regression and classification components. The losses are defined as follows:
\begin{equation}
\begin{split}
\mathcal{L}_{det} &= \lambda_{det\_reg}\mathcal{L}_{det\_reg} + \lambda_{det\_cls}\mathcal{L}_{det\_cls}, \\
\mathcal{L}_{map} &= \lambda_{map\_reg}\mathcal{L}_{map\_reg} + \lambda_{map\_cls}\mathcal{L}_{map\_cls}, \\
\mathcal{L}_{mot} &= \lambda_{mot\_reg}\mathcal{L}_{mot\_reg} + \lambda_{mot\_cls}\mathcal{L}_{mot\_cls}, \\
\mathcal{L}_{plan} &= \lambda_{plan\_reg}\mathcal{L}_{plan\_reg} + \lambda_{plan\_cls}\mathcal{L}_{plan\_cls}, \\
\mathcal{L}_{total} &= \mathcal{L}_{det}+\mathcal{L}_{map}+\mathcal{L}_{mot}+\mathcal{L}_{plan}.
\end{split}
\end{equation}

The loss weights are set as follows: $\lambda_{det\_reg}=0.25$, $\lambda_{det\_cls}=2.0$, $\lambda_{map\_reg}=10.0$, $\lambda_{map\_cls}=1.0$, $\lambda_{mot\_reg}=0.05$, 
$\lambda_{mot\_cls}=0.1$, 
$\lambda_{plan\_reg}=1.0$, 
$\lambda_{plan\_cls}=0.5$.

\subsection{Notations}
As shown in Table~\ref{tab:supp_notation}, we provide a lookup table for notations used in the paper.

\begin{table*} [ht!] 
    \centering
    {\begin{tabular}{c|l}
       \toprule[1.5pt]
        \textbf{Notation} & \textbf{Description} \\\midrule
        
        $N_{\rm a}$ & the number of surrounding agents \\
        $M_{\rm mot}$ & the number of prediction modes \\
        $C$ & the feature channels \\
        $T_{\rm mot}$ & the number of future time steps for prediction \\
        $M_{\rm plan}$ & the number of planning modes \\
        $T_{\rm plan}$ & the number of future time steps for planning \\
        $K$ & the number of historical motion planning frames stored in the memory queue \\
        $N_{\rm img}$ & the number of camera views \\
        $\mathcal{F}$ & multi-view visual features \\
        $Q_{\rm obj}$ & object queries \\
        $B_{\rm obj}$ & object anchor boxes \\
        $Q_{\rm mot}$ & motion queries \\
        $Q_{\rm plan}$ & planning queries \\
        $Q_{\rm m2d}$ & historical motion queries used in the Historical Mot2Det Fusion Module \\
        $T_{\rm m2m}$ & the number of time steps that interact with historical motion queries \\
        $Q_{\rm m2m}$ & historical motion queries used in the History-Enhanced Motion Prediction Module \\
        $T_{\rm p2p}$ & the number of time steps that interact with historical planning queries \\
        $Q_{\rm p2p}$ & historical planning queries used in the History-Enhanced Planning Module \\
        $Q_{\rm mot}^{*}$ & selected motion queries used in the Step-Level Mot2Plan Interaction Module \\
        
        \bottomrule[1.5pt]
    \end{tabular}}
    \caption{Notations used in the paper.}
    \label{tab:supp_notation}
\end{table*}

\section{Experiments}
\subsection{Evaluation metrics}

\paragraph{Open-loop evaluation.}
We provide evaluation metrics for perception, prediction, and planning tasks. 
The detection and tracking evaluation adheres to standard protocols~\cite{nuscenes}. 
For detection, we use mean Average Precision (\textbf{mAP}) and nuScenes Detection Score (\textbf{NDS}).
For tracking, Average Multi-object Tracking Accuracy (\textbf{AMOTA}), Average Multi-object Tracking Precision (\textbf{AMOTP}), and Identity Switches (\textbf{IDS}).
The online mapping~\cite{sparsedrive} and motion prediction~\cite{UniAD,sparsedrive} evaluations are consistent with previous works.
For online mapping, we use the Average Precision (\textbf{AP}) for three map classes: lane divider, pedestrian crossing, and road boundary. The mean Average Precision (\textbf{mAP}) is then calculated by averaging the AP across all classes.
For motion prediction, we use the minimum Average Displacement Error (\textbf{ADE}), minimum Final Displacement Error (\textbf{FDE}), Miss Rate (\textbf{MR}), and End-to-End Prediction Accuracy (\textbf{EPA}) as proposed in ViP3D~\cite{vip3d}.
For planning, we use the L2 Displacement Error metric, as used in VAD~\cite{vad}, and the Collision Rate, as defined in~\cite{bevplanner,sparsedrive}.
The Collision Rate addresses two issues in the previous benchmark~\cite{UniAD,vad}: false collisions in certain cases and the exclusion of the ego vehicle's heading.

\paragraph{Closed-loop evaluation on NeuroNCAP.} Following the official definition~\cite{neuroncap}, a NeuroNCAP score is computed for each scenario. A full score is awarded only if a collision is completely avoided, while partial scores are granted for successfully reducing impact velocity. Inspired by the 5-star Euro NCAP rating system~\cite{euroNCAP2023collision}, the NeuroNCAP score is calculated as:
\begin{equation}
\label{eq:ncap-score}
    \text{NNS} = 
    \begin{cases}
        5.0                                    & \text{if no collision}, \\
        4.0 \cdot \text{max}(0, 1 - v_i / v_r) & \text{otherwise}.
    \end{cases}\enspace
\end{equation}
where $v_i$ is the impact speed as the magnitude of relative velocity between ego-vehicle and colliding actor, and $v_r$ is the reference impact speed that would occur if no action is performed. In other words, the score corresponds to a 5-star rating if collision is entirely avoided, and otherwise the rating is linearly decreased from four to zero stars at (or exceeding) the reference impact speed.

\subsection{Implementation details}
As stated in the Implementation Details section of the main paper, training is conducted in two stages. The first stage focuses on the perception task with a batch size of 8 for 100 epochs, while the second stage focuses on end-to-end training with a batch size of 4 for 15 epochs. The total training time is approximately 1.5 days.
For the model settings, the number of object queries and map queries is set to 900 and 100, respectively. The feature dimension $C$ is 256. The backbone, ResNet101, uses pre-trained weights from the nuImage dataset.

\subsection{Online mapping results}
The online mapping results on the nuScenes~\cite{nuscenes} validation dataset, compared to other methods, are shown in Table~\ref{tab:supp_map}.

\begin{table} [ht!] 
    \centering
    \setlength{\tabcolsep}{0.6mm}

        \centering
        \resizebox{1.0\columnwidth}{!}
        {\begin{tabular}[b]{l|ccc|c}
        \toprule[1.5pt]

        \textbf{Method} & \textbf{AP$_{ped}$} $\uparrow$ & \textbf{AP$_{divider}$} $\uparrow$ & \textbf{AP$_{boundary}$} $\uparrow$ & \textbf{mAP} $\uparrow$ \\ 
        \midrule 
         
        HDMapNet~\cite{hdmapnet} & 14.4 & 21.7 & 33.0 & 23.0  \\
        VectorMapNet~\cite{vectormapnet} & 36.1 & 47.3 & 39.3 & 40.9  \\ 
        MapTR~\cite{MapTR} & 56.2 & 59.8 & 60.1 & 58.7  \\\midrule 
        VAD$^{\dag}$~\cite{vad} & 40.6 & 51.5 & 50.6 & 47.6   \\ 
        SparseDrive~\cite{sparsedrive}  & 49.9 & 57.0 & 58.4 & 55.1  \\

        \rowcolor{gray!20}
        \netName-S~(Ours)~& 51.8 & 56.4 & 57.5 & 55.2 \\
        \rowcolor{gray!20}
        \netName-B~(Ours)~& 52.0 & 57.1 & 57.9 & 55.7 \\
        \bottomrule[1.5pt]
        \end{tabular}}
    
    \caption{Comparison of online mapping results for state-of-the-art online mapping and end-to-end methods. $\dag$ indicates evaluation with the official checkpoint.}
    \label{tab:supp_map}
\end{table}

\subsection{Comparison with other baselines}
We compare our model with two other common methods, and the results are shown in Table~\ref{tab:re_baseline}.

\begin{table} [ht!] 
    \centering
    \setlength{\tabcolsep}{0.5mm}

        \centering
        \resizebox{1.0\columnwidth}{!}
        {\begin{tabular}[b]{l|cccc|cccc}
        \toprule[1.5pt]

        \multirow{2}{*}{\textbf{Method}} & 
        \multicolumn{4}{c|}{\textbf{L2 ($m$)} $\downarrow$} & 
        \multicolumn{4}{c}{\textbf{Col. Rate (\%)} $\downarrow$}  \\
        & 1$s$ & 2$s$ & 3$s$ & Avg. & 1$s$ & 2$s$ & 3$s$ & Avg. \\
        \midrule 
         
         BEVPlanner~\cite{bevplanner} & 0.27 & 0.54 & 0.90 & 0.57 & 0.04 & 0.35 & 1.80 & 0.73 \\
         BEVPlanner*~\cite{bevplanner} & 0.28 & 0.42 & 0.68 & 0.46 & 0.04 & 0.37 & 1.07 & 0.49 \\
         PARA-Drive*~\cite{PARAdrive} & 0.25 & 0.46 & 0.74 & 0.48 & 0.14 & 0.23 & 0.39 & 0.25 \\
         \rowcolor{gray!20}
         \netName & 0.29 & 0.57 & 0.92 & 0.59 & 0.01 & 0.05 & 0.22 & 0.09 \\
         
         \bottomrule[1.5pt]
        \end{tabular}}
    
    \caption{Comparison with other baselines. ``*" denotes use ego status as input.}
    \label{tab:re_baseline}
\end{table}

\subsection{Analysis for moving agents}
Following the reviewer’s suggestion, we evaluate our model using a more suitable metric proposed by~\cite{reICRA}, which better reflects the end-to-end nature of the task (see Table~\ref{tab:re_eval}). As shown, our model outperforms UniAD and ViP3D on these metrics, which specifically focus on moving agents.

\begin{table} [ht!] 
    \centering
    \setlength{\tabcolsep}{1.4mm}

        \centering

        {\begin{tabular}[b]{l|cccc}
        \toprule[1.5pt]
        
        \textbf{Method} & \textbf{mAP${_f}$} $\uparrow$ & \textbf{minADE} $\downarrow$  & \textbf{minFDE} $\downarrow$  & \textbf{MR} $\downarrow$ \\\midrule
         
        ViP3D~\cite{vip3d} & 0.034 & 3.540 & 5.943 & 0.432 \\
        UniAD~\cite{UniAD} & 0.117 & 1.842 & 3.258 & 0.228 \\
        \rowcolor{gray!20}
        \netName & 0.139 & 1.733 & 3.098 & 0.210 \\
         
         \bottomrule[1.5pt]
        \end{tabular}}
    
    \caption{Motion forecasting results with more adapted metrics. We use 6 modes by default.}
    \label{tab:re_eval}
\end{table}


\subsection{Safety assessments}
Following the reviewer’s suggestion, we conduct a safety assessment of our method, including an analysis of its robustness to images, as shown in Table~\ref{tab:re_safety}. Additionally, we provide an analysis of failure cases and limitations in the supplementary material.

\begin{table} [ht!] 
    \centering
    \setlength{\tabcolsep}{0.7mm}

        \centering
        \resizebox{1.0\columnwidth}{!}
        {\begin{tabular}[b]{c|c|c}
        \toprule[1.5pt]

        \textbf{Image Corruption} & 
        \textbf{L2 ($m$)} $\downarrow$ Avg. & 
        \textbf{Col. Rate (\%)} $\downarrow$ Avg. \\
        \midrule 
         Only front view & 0.68 & 0.22 \\
         Blank & 2.76 & 1.83 \\
         \rowcolor{gray!20}
         Default & 0.59 & 0.09\\
         
         \bottomrule[1.5pt]
        \end{tabular}}
    
    \caption{Our model’s robustness to images on nuScenes.}
    \label{tab:re_safety}
\end{table}

\subsection{Experiments on the Bench2Drive dataset}
We conduct experiments on CARLA v2 simulator using the Bench2Drive benchmark~\cite{Bench2Drive}, as shown in Table~\ref{tab:bench2drive}. Our method outperforms UniAD and VAD in both open-loop and closed-loop evaluations, showcasing the model's generalization ability.

\begin{table} [ht!]
\centering

{\begin{tabular}{l|c|cc}
\toprule[1.5pt]

\multirow{2}{*}{\textbf{Method}} & \multicolumn{1}{c|}{\textbf{Open-loop}} & \multicolumn{2}{c}{\textbf{Closed-loop}} \\ \cmidrule{2-4}
 & \multicolumn{1}{c|}{Avg. L2 $\downarrow$} & \multicolumn{1}{c}{DS $\uparrow$}  & \multicolumn{1}{c}{SR (\%) $\uparrow$} \\ \midrule
 
 AD-MLP~\cite{admlp} & \multicolumn{1}{c|}{3.64} & \multicolumn{1}{c}{18.05} & \multicolumn{1}{c}{0.00} \\ 
 
 UniAD~\cite{UniAD} & \multicolumn{1}{c|}{0.73} & \multicolumn{1}{c}{45.81} & \multicolumn{1}{c}{16.36} \\ 
 
 VAD~\cite{vad} & \multicolumn{1}{c|}{0.91} & \multicolumn{1}{c}{42.35} & \multicolumn{1}{c}{15.00} \\ 

 \rowcolor{gray!20}

 \netName & \multicolumn{1}{c|}{0.71} & \multicolumn{1}{c}{50.06} & \multicolumn{1}{c}{22.73} \\ 

 \midrule

 TCP*~\cite{TCP} & \multicolumn{1}{c|}{1.70} & \multicolumn{1}{c}{40.70} & \multicolumn{1}{c}{15.00} \\
 TCP-ctrl*~\cite{TCP} & \multicolumn{1}{c|}{-} & \multicolumn{1}{c}{30.47} & \multicolumn{1}{c}{7.27} \\
 TCP-traj*~\cite{TCP} & \multicolumn{1}{c|}{1.70} & \multicolumn{1}{c}{59.90} & \multicolumn{1}{c}{30.00} \\
 ThinkTwice*~\cite{thinktwice} & \multicolumn{1}{c|}{0.95} & \multicolumn{1}{c}{62.44} & \multicolumn{1}{c}{31.23} \\
 DriveAdapter*~\cite{driveadapter} & \multicolumn{1}{c|}{1.01} & \multicolumn{1}{c}{64.22} & \multicolumn{1}{c}{33.08} \\
 
\bottomrule[1.5pt]
 
\end{tabular}}

\caption{Experiment on CARLA v2 using the Bench2Drive benchmark. ``DS" indicates Driving Score, ``SR" indicates Success Rate. ``*" denotes expert feature distillation.} 
\label{tab:bench2drive}

\end{table}

\subsection{Ablation study}
\paragraph{Effects of self-attention in motion prediction.}
We conduct an ablation study to evaluate the effects of step-level and mode-level self-attention in the motion prediction module, as shown in Table~\ref{tab:supp_attnmot}, similar to Table 7 in the main paper. Both types of self-attention propagate historical information across prediction steps and modes, enhancing the accuracy of motion prediction.

\begin{table} [ht!] 
    \centering
    \setlength{\tabcolsep}{2.0mm}

        \centering
        {\begin{tabular}[b]{cc|c
        |c|c}
        \toprule[1.5pt]

        \multirow{2}{*}{\textbf{SLA}} & \multirow{2}{*}{\textbf{MLA}} & \textbf{ADE~($m$)} $\downarrow$ & \textbf{FDE~($m$)} $\downarrow$ & \textbf{EPA} $\uparrow$ \\ 
        & & Car / Ped & Car / Ped & Car / Ped\\\midrule 
         
         \ding{51} & & 0.65 / 0.71 & 1.02 / 1.00 & 0.49 / 0.42 \\
         & \ding{51} & 0.64 / 0.71 & 1.00 / 1.01 & 0.48 / 0.42 \\ 
         \rowcolor{gray!20}
         \ding{51} & \ding{51} & 0.62 / 0.70 & 0.98 / 0.99 & 0.50 / 0.44 \\
         
         \bottomrule[1.5pt]
        \end{tabular}}
    
    \caption{Ablation study on step-level self-attention (SLA) and mode-level self-attention (MLA).}
    \label{tab:supp_attnmot}
\end{table}

\paragraph{Effects of the number of historical frames.}
We conduct an ablation study on the number of historical frames $K$, as shown in Table~\ref{tab:supp_k}. The results show that $K=3$ achieves the best balance between efficiency and performance.

\begin{table} [ht!] 
    \centering

        \centering
        {\begin{tabular}[b]{c|c|c}
        \toprule[1.5pt]

        \textbf{HisFrame} &
        Avg. \textbf{L2 ($m$)} $\downarrow$  & 
        Avg. \textbf{Col. Rate (\%)} $\downarrow$ \\
        \midrule 
        
         2 & 0.64 & 0.13 \\
         
         \rowcolor{gray!20}
         3 & 0.59 & 0.09 \\
         
         4 & 0.62 & 0.10 \\
         \bottomrule[1.5pt]
        \end{tabular}}
    
    \caption{Ablation study on the number of historical frames.}
    \label{tab:supp_k}
\end{table}

\subsection{Qualitative results}
\begin{figure*}[ht!]
    \centering
    
        \includegraphics[width=1\textwidth]
        {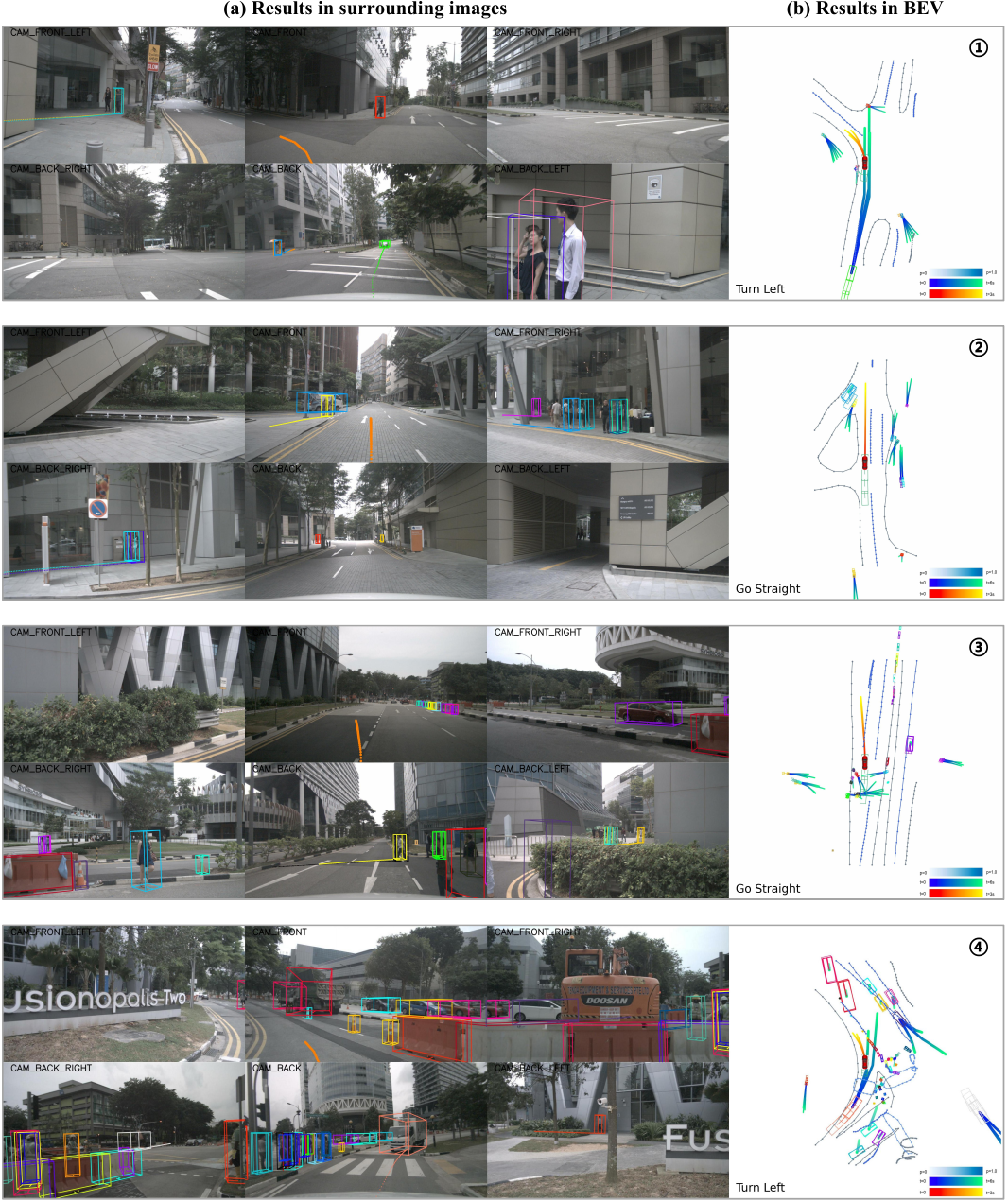}
        
    \caption{
    Qualitative results in the \textbf{open-loop} evaluation.
    }
    \label{fig:supp_open}
\end{figure*}

\begin{figure*}[ht!]
    \centering
    
        \includegraphics[width=1\textwidth]
        {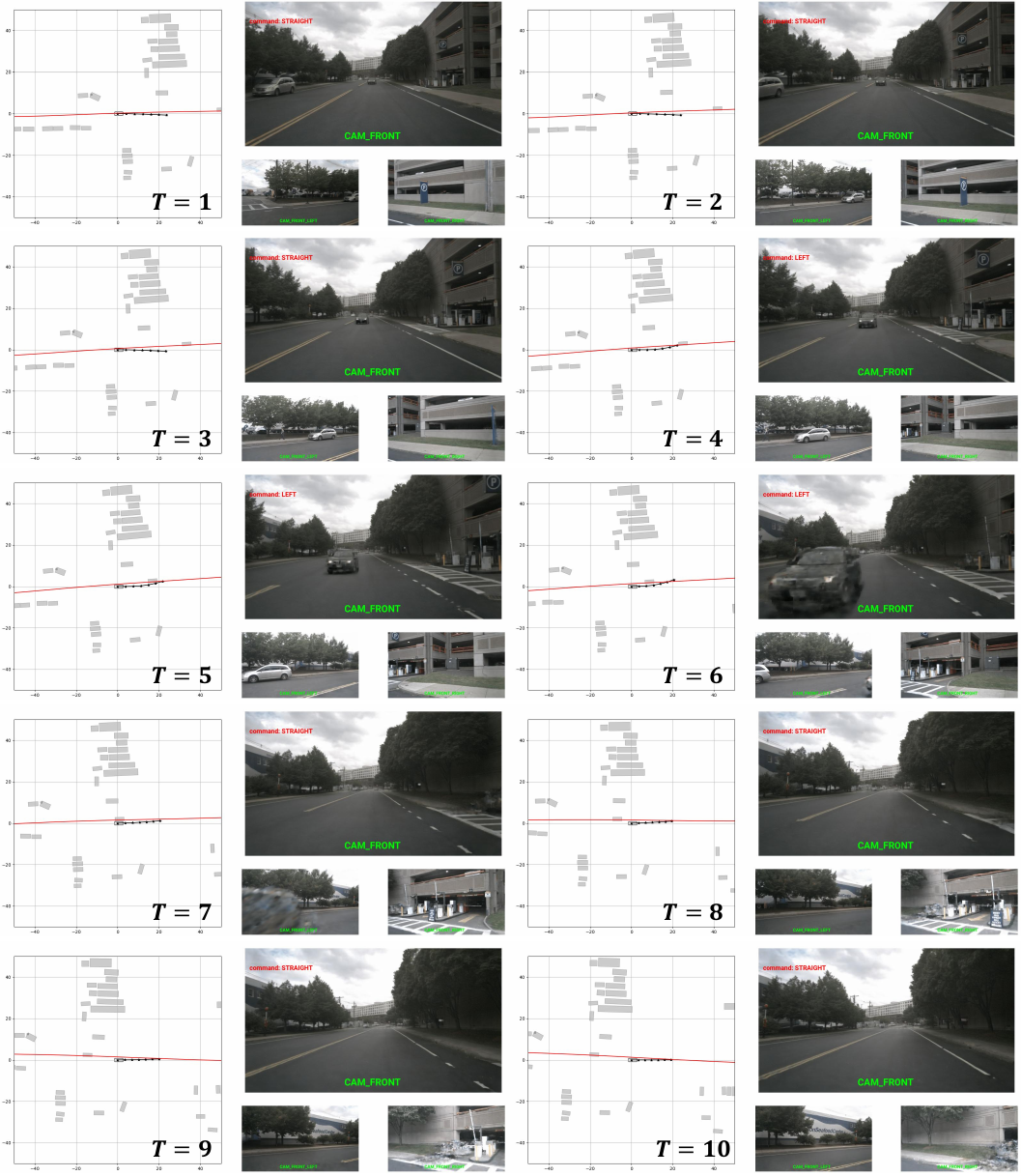}
        
    \caption{
    Qualitative result 1 in the \textbf{closed-loop} evaluation.
    }
    \label{fig:supp_close1}
\end{figure*}

\begin{figure*}[ht!]
    \centering
    
        \includegraphics[width=1\textwidth]
        {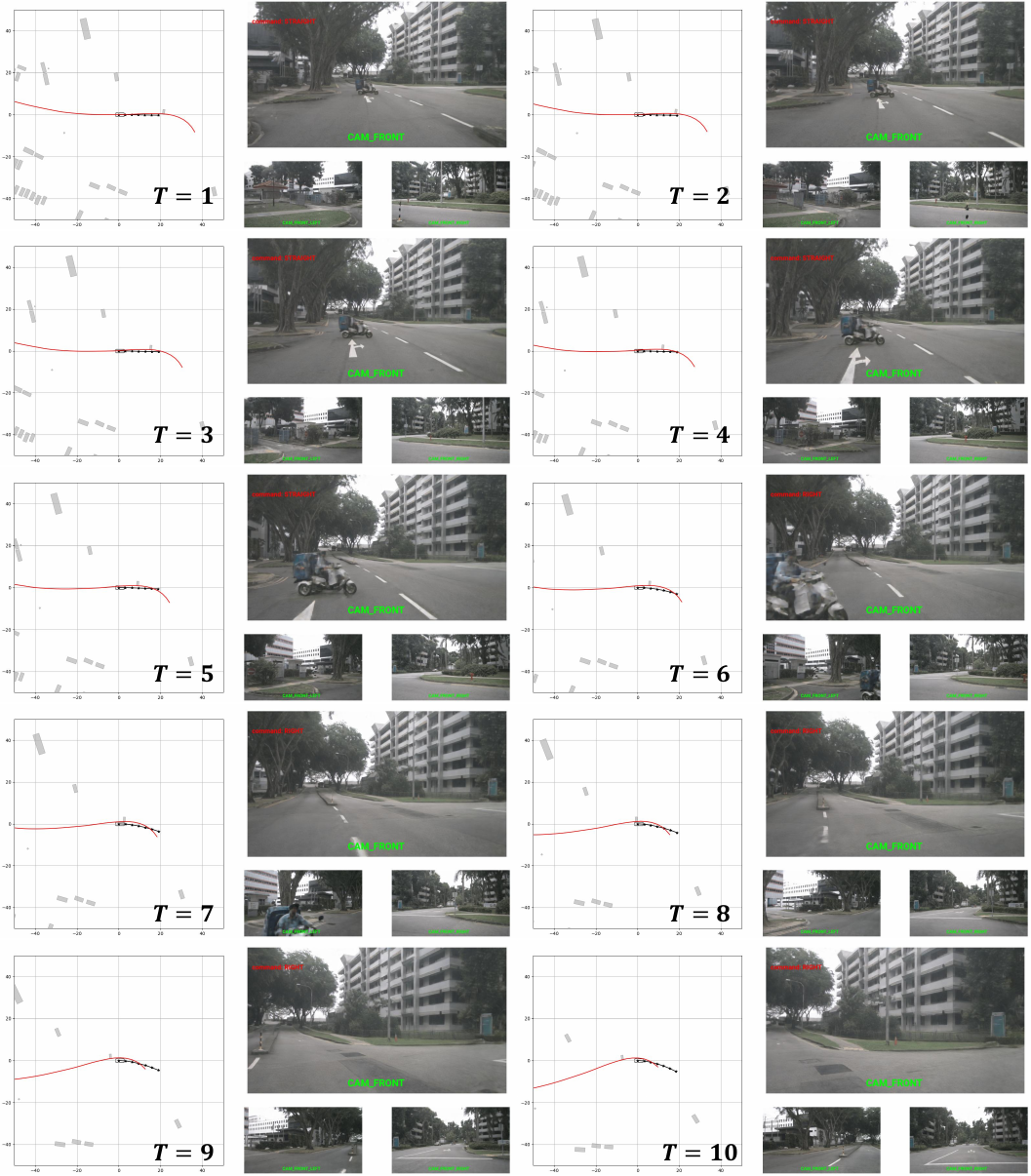}
        
    \caption{
    Qualitative result 2 in the \textbf{closed-loop} evaluation.
    }
    \label{fig:supp_close2}
\end{figure*}

\begin{figure*}[ht!]
    \centering
    
        \includegraphics[width=1\textwidth]
        {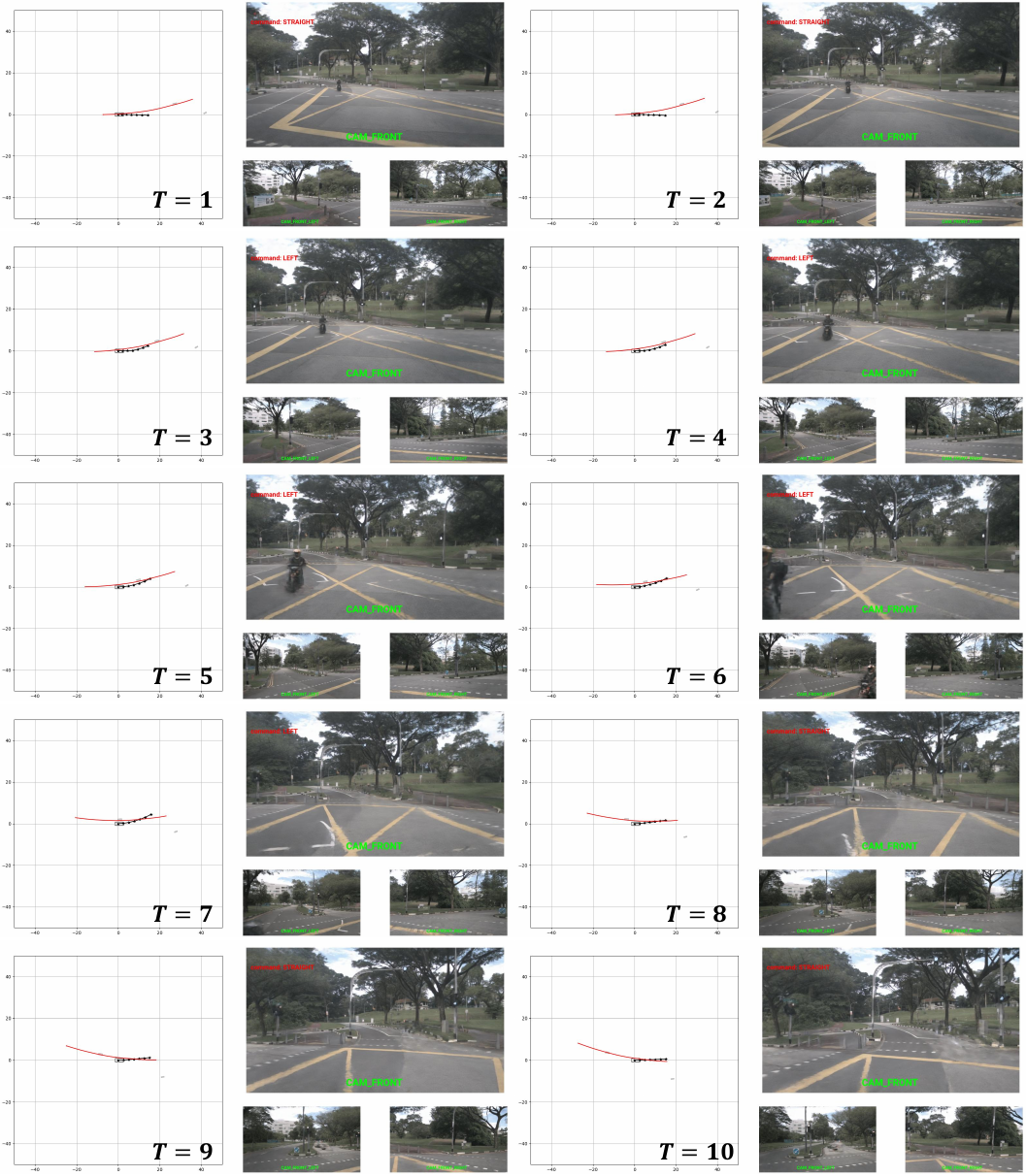}
        
    \caption{
    Qualitative result 3 in the \textbf{closed-loop} evaluation.
    }
    \label{fig:supp_close3}
\end{figure*}

We present additional qualitative results from both the open-loop and closed-loop evaluations on the nuScenes~\cite{nuscenes} dataset. The open-loop evaluation results are shown in Figure~\ref{fig:supp_open}. The closed-loop evaluation results, obtained using the NeuroNCAP~\cite{neuroncap} simulator, are shown in Figures~\ref{fig:supp_close1}, \ref{fig:supp_close2}, and \ref{fig:supp_close3}. Notably, the red line in the closed-loop evaluation represents the reference trajectory under normal driving conditions, where no safety risk is present.

\subsection{Failure cases}
We present the failure cases observed in both open-loop and closed-loop evaluations. 

\begin{figure*}[ht!]
    \centering
    
        \includegraphics[width=1\textwidth]
        {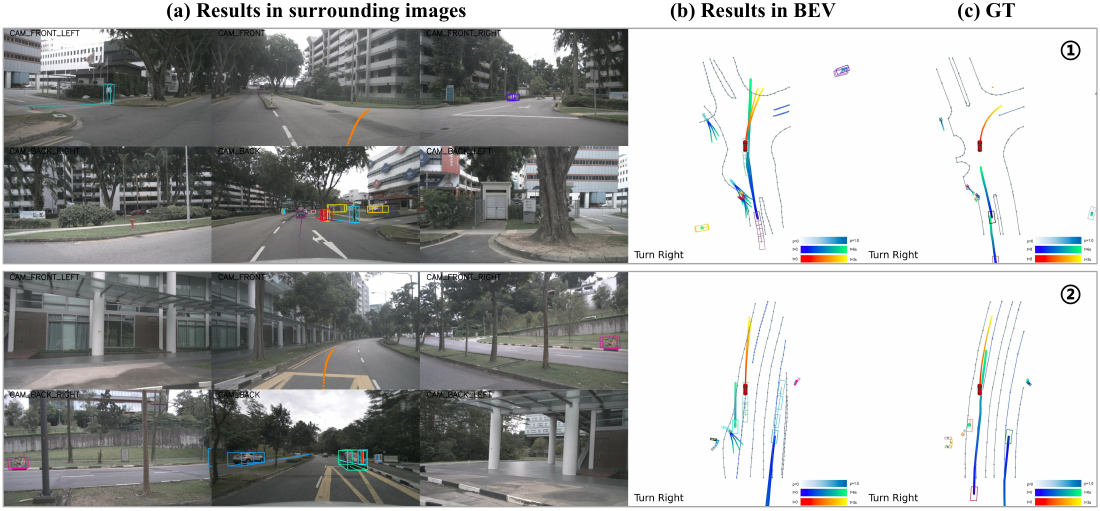}
        
    \caption{
    Failure cases in the \textbf{open-loop} evaluation.
    }
    \label{fig:supp_fial}
\end{figure*}

The failure cases from the open-loop evaluation are shown in Figure~\ref{fig:supp_fial}. In both the first and second cases, the planned trajectories veer off the road at the curbs (road boundaries). Adding constraints or post-processing techniques to keep the planned trajectories on the road could prevent these failures.

\begin{figure*}[ht!]
    \centering
    
        \includegraphics[width=1\textwidth]
        {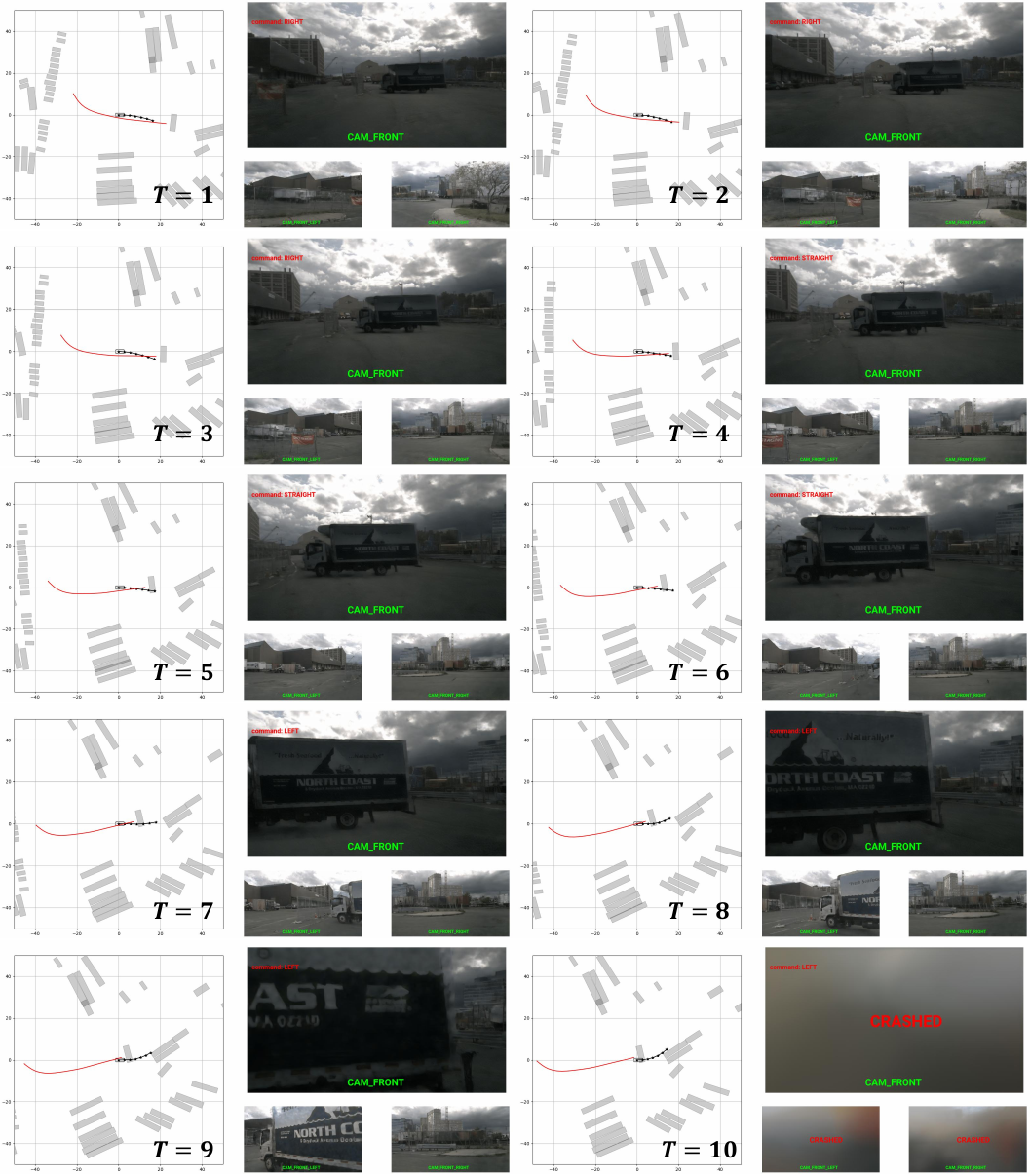}
        
    \caption{
    Failure case in the \textbf{closed-loop} evaluation.
    }
    \label{fig:supp_fialclose}
\end{figure*}

The failure case from the closed-loop evaluation is shown in Figure~\ref{fig:supp_fialclose}. The planned trajectories steer to avoid the front truck, but insufficient steering and a lack of deceleration still result in a crash. Providing more training data focused on deceleration or applying post-processing techniques to enforce slowing down could prevent this failure.

\section{Limitations and future work}
The results of closed-loop testing indicate that our model still struggles to handle safety-critical scenarios and relies heavily on complex post-processing. This limitation is a common issue among existing end-to-end methods. Our approach mitigates safety-critical scenarios to some extent by aggregating historical planning information to produce coherent driving actions that avoid collisions. However, this remains insufficient. Exploring effective and efficient solutions, such as training with more data in these situations or integrating the end-to-end pipeline with reinforcement learning or rule-based planning, is a promising direction for future research.

\section{Discussion}

\subsection{Further explanation about our~\netName}

\begin{figure}[t!]
    \centering
    
        \includegraphics[width=0.35\textwidth]{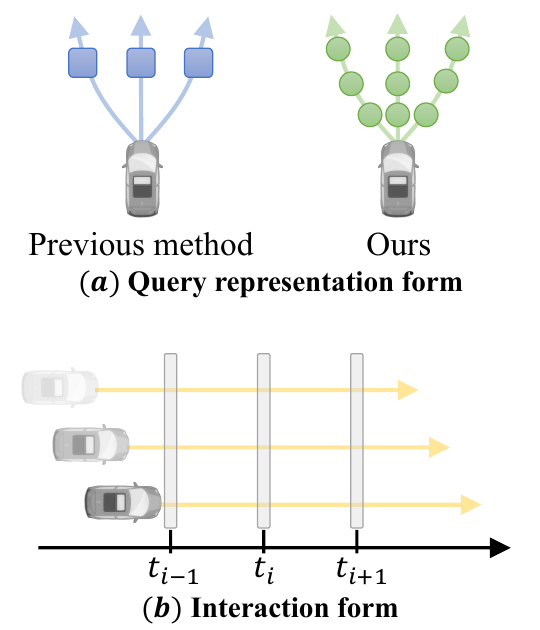}
        
    \caption{
    Further explanation about our~\netName.
    }
    \label{fig:supp_eg}
\end{figure}

To better illustrate our method, we provide a further explanation of our key idea. As shown in Figure~\ref{fig:supp_eg} (a), unlike previous methods~\cite{UniAD,vad,sparsedrive}, we represent motion and planning queries as multi-step queries. In contrast to previous approaches that use a single query to represent an entire trajectory instance, our method utilizes multiple queries for a single trajectory. For example, in the planning task on the nuScenes dataset, where a 3-second future trajectory is planned at 2 Hz, six queries are used to represent one trajectory instance.

Regarding the interaction mechanism in our method, as shown in Figure~\ref{fig:supp_eg} (b), queries are grouped based on time steps, and those corresponding to the same time step interact through our designed modules. This approach is applied to both motion queries for surrounding agents and planning queries for the ego agent.

\subsection{Discussion about belief states}

Belief states represent an agent's probabilistic estimation of the true state of the environment, given past observations and actions. They are commonly used in decision-making under uncertainty, where the full state is not directly observable. By maintaining and updating a belief state, an agent can make more informed and robust decisions in dynamic or partially observable environments. 
Some methods~\cite{gu2021belief,bouton2017belief,huang2024learning} explore its potential for planning and decision-making in autonomous driving.
Huang \etal~\cite{huang2024learning} proposes a neural memory-based belief update model for online behavior prediction and a macro-action-based MCTS planner guided by deep Q-learning. By leveraging long-term multi-modal trajectory predictions and optimizing decision-making under uncertainty, the approach enhances both efficiency and performance in autonomous driving scenarios. 

Our~\netName~can essentially be seen as encoding belief states. By leveraging historical prediction and planning, it incorporates belief states into perception, prediction, and planning, enhancing end-to-end autonomous driving performance.

\subsection{Discussion about historical predictions}

In the motion prediction task, recent works have explored leveraging historical predictions to improve performance. HPNet~\cite{hpnet} utilizes historical predictions to achieve more stable and accurate motion forecasts, while RealMotion~\cite{RealMotion} operates in a streaming fashion to enhance motion prediction. In contrast, our~\netName~incorporates both historical prediction and planning to optimize the entire pipeline of end-to-end autonomous driving.

\newpage



\end{document}